\documentclass[11pt,a4paper]{article}
\usepackage[hyperref]{acl2020}
\usepackage{times}
\usepackage{latexsym}
\usepackage{amssymb}
\usepackage{amsmath}

\usepackage[pdftex]{graphicx}

\usepackage{subcaption}
\usepackage{microtype}

\aclfinalcopy 
\def\aclpaperid{394} 

\usepackage{array}
\newcolumntype{P}[1]{>{\centering\arraybackslash}p{#1}}

\title{Reducing Gender Bias in Neural Machine Translation as a Domain Adaptation Problem}
  
\author{Danielle Saunders \and Bill Byrne \\
    Department of Engineering, University of Cambridge, UK  \\
      {\tt \{ds636, wjb31\}@cam.ac.uk}}

\begin{document}
\maketitle
\begin{abstract}
Training data for NLP tasks often exhibits gender bias in that fewer sentences refer to women than to men. In Neural Machine Translation (NMT) gender bias has been shown to reduce translation quality, particularly when the target language has grammatical gender. The recent WinoMT challenge set allows us to measure this effect directly \citep{stanovsky-etal-2019-evaluating}.

Ideally we would reduce system bias by simply debiasing all data prior to training, but achieving this effectively is itself a challenge. Rather than attempt to create a `balanced' dataset, we use transfer learning on a small set of trusted, gender-balanced examples. This approach gives strong and consistent improvements in gender debiasing with much less computational cost than training from scratch. 

A known pitfall of transfer learning on new domains is `catastrophic forgetting', which we address both in adaptation and in inference. During adaptation we show that Elastic Weight Consolidation allows a performance trade-off between general translation quality and bias reduction. During inference we propose a lattice-rescoring scheme which outperforms all systems evaluated in \citet{stanovsky-etal-2019-evaluating} on WinoMT with no degradation of general test set BLEU, and we show this scheme can be applied to remove gender bias in the output of `black box` online commercial MT systems. We demonstrate our approach translating from English into three languages with varied linguistic properties and data availability.

\end{abstract}

\section{Introduction}
As language processing tools become more prevalent concern has grown over their susceptibility to social biases and their potential to propagate bias \citep{hovy-spruit-2016-social, sun-etal-2019-mitigating}. Natural language training data inevitably reflects biases present in our society. For example, gender bias manifests itself in training data which features more examples of men than of women. Tools trained on such data will then exhibit or amplify the biases \citep{zhao-etal-2017-men} and their harmful stereotypes.

Gender bias is a particularly important problem for Neural Machine Translation (NMT) into gender-inflected languages. An over-prevalence of some gendered forms in the training data leads to translations with identifiable errors \citep{stanovsky-etal-2019-evaluating}. Translations are better for sentences involving men and for sentences containing stereotypical gender roles. For example, mentions of male doctors are more reliably translated than those of male nurses  \citep{sun-etal-2019-mitigating, prates2019assessing}.

Recent approaches to the bias problem in NLP have involved training from scratch on artificially gender-balanced versions of the original dataset \citep{zhao-etal-2018-gender, zmigrod-etal-2019-counterfactual} or with de-biased embeddings \citep{escude-font-costa-jussa-2019-equalizing, bolukbasi2016man}. While these approaches may be effective, training from scratch is inefficient and gender-balancing embeddings or large parallel datasets are challenging problems \citep{gonen2019lipstick}. 

Instead we propose treating gender debiasing as a domain adaptation problem, since NMT models can very quickly adapt to a new domain \citep{freitag2016fast}. To the best of our knowledge this work is the first to attempt NMT bias reduction by fine-tuning, rather than retraining. We consider three aspects of this adaptation problem: creating less biased adaptation data, parameter adaptation using this data, and inference with the debiased models produced by adaptation.

Regarding data, we suggest that a small, trusted gender-balanced set could allow more efficient and effective gender debiasing than a larger, noisier set. To explore this we create a tiny, handcrafted profession-based dataset for transfer learning. For contrast, we also consider fine-tuning on a counterfactual subset of the full dataset and propose a straightforward scheme for artificially gender-balancing parallel text for NMT. 

We find that during domain adaptation improvement on the gender-debiased domain comes at the expense of translation quality due to catastrophic forgetting \citep{french1999catastrophic}. We can balance improvement and forgetting with a regularised training procedure, Elastic Weight Consolidation (EWC), or in inference by a two-step lattice rescoring procedure.

We experiment with three language pairs, assessing the impact of debiasing on general domain BLEU and on the WinoMT challenge set \citep{stanovsky-etal-2019-evaluating}. We find that continued training on the handcrafted set gives far stronger and more consistent improvements in gender-debiasing with orders of magnitude less training time, although as expected general translation performance as measured by BLEU decreases. 

We further show that regularised adaptation with EWC can reduce bias while limiting degradation in general translation quality. We also present a lattice rescoring procedure in which initial hypotheses produced by the biased baseline system are transduced to create gender-inflected search spaces which can be rescored by the adapted model. We believe this approach, rescoring with models targeted to remove bias, is novel in NMT. The rescoring procedure improves WinoMT accuracy by up to 30\%  with no decrease in BLEU on the general test set.

Recent recommendations for ethics in Artificial Intelligence have suggested that social biases or imbalances in a dataset be addressed prior to model training \citep{hleg2019ethics}. This recommendation presupposes that the source of bias in a dataset is both obvious and easily adjusted. We show that debiasing a full NMT dataset is difficult, and suggest alternative efficient and effective approaches for debiasing a model after it is trained. This avoids the need to identify and remove all possible biases prior to training, and has the added benefit of preserving privacy, since no access to the original data or knowledge of its contents is required.  As evidence,  in section \ref{sec:commercial}, we show this scheme can be applied to remove gender bias in the output of ‘black box‘ online commercial MT systems.

\subsection{Related work}
\citet{vanmassenhove-etal-2018-getting} treat gender as a domain for machine translation, training from scratch by augmenting Europarl data with a tag indicating the speaker's gender. This does not inherently remove gender bias from the system but allows control over the translation hypothesis gender. \citet{moryossef-etal-2019-filling} similarly prepend a short phrase at inference time which acts as a gender domain label for the entire sentence. These approaches are not directly applicable to text which may have more than one gendered entity per sentence, as in coreference resolution tasks.

\citet{escude-font-costa-jussa-2019-equalizing} train NMT models from scratch with debiased word embeddings. They demonstrate improved performance on an English-Spanish occupations task with a single profession and pronoun per sentence. We assess our fine-tuning approaches on the WinoMT coreference set, with two entities to resolve per sentence. 

For monolingual NLP tasks a typical approach is gender debiasing using counterfactual data augmentation where for each gendered sentence in the data a gender-swapped equivalent is added. \citet{zhao-etal-2018-gender} show improvement in coreference resolution for English using counterfactual data. \citet{zmigrod-etal-2019-counterfactual} demonstrate a more complicated scheme for gender-inflected languages. However, their system focuses on words in isolation, and is difficult to apply to co-reference and conjunction situations with more than one term to swap, reducing its practicality for large MT datasets.

Recent work recognizes that NMT can be adapted to domains with desired attributes using small datasets \citep{farajian-etal-2017-multi, michel-neubig-2018-extreme}. Our choice of a small, trusted dataset for adaptation specifically to a debiased domain connects also to recent work in data selection by \citet{wang-etal-2018-denoising}, in which fine-tuning on less noisy data reduces translation noise. Similarly we propose fine-tuning on less biased data to reduce gender bias in translations. This is loosely the inverse of the approach described by \citet{park-etal-2018-reducing} for monolingual  abusive language detection, which pre-trains on a larger, less biased set. 

\section{Gender bias in machine translation}
We focus on translating coreference sentences containing professions as a representative subset of the gender bias problem. This follows much recent work on NLP gender bias \citep{rudinger-etal-2018-gender, zhao-etal-2018-gender, zmigrod-etal-2019-counterfactual} including the release of WinoMT, a relevant challenge set for NMT \citep{stanovsky-etal-2019-evaluating}. 

A sentence that highlights gender bias is:

\textit{The \textbf{doctor} told the nurse that \textbf{she} had been busy.}

A human translator carrying out coreference resolution would infer that `she' refers to the doctor, and correctly translate the entity to German as \textit{Die \"Arztin}. An NMT model trained on a biased dataset in which most doctors are male might incorrectly default to the masculine form, \textit{Der Arzt}.

Data bias does not just affect translations of the stereotyped roles. Since NMT inference is usually left-to-right, a mistranslation can lead to further, more obvious mistakes later in the translation. For example, our baseline en-de system translates the English sentence

\textit{The cleaner hates the \textbf{developer} because \textbf{she} always leaves the room dirty.}

to the German

\textit{Der Reiniger ha{\ss}t  \textbf{den Entwickler}, weil \textbf{er} den Raum immer schmutzig l{\"a}sst.}

Here not only is `developer' mistranslated as the masculine \textit{den Entwickler} instead of the feminine \textit{die Entwicklerin}, but an unambiguous pronoun translation later in the sentence is incorrect: \textit{er} (`he') is produced instead of \textit{sie} (`she').

In practice, not all translations with gender-inflected words can be unambiguously resolved. A simple example is:

\textit{The doctor had been busy.}

This would likely be translated with a masculine entity according to the conventions of a language, unless extra-sentential context was available. As well, some languages have adopted gender-neutral singular pronouns and profession terms, both to include non-binary people and to avoid the social biases of gendered language \citep{misersky2019grammatical}. However, the target languages supported by WinoMT lack widely-accepted non-binary inflection conventions \citep{ackerman2019syntactic}. This paper addresses gender bias that can be resolved at the sentence level and evaluated with existing test sets, and does not address these broader challenges.

\subsection{WinoMT challenge set and metrics}
WinoMT \citep{stanovsky-etal-2019-evaluating} is a recently proposed challenge set for gender bias in NMT. Moreover it is the only significant challenge set we are aware of to evaluate translation gender bias comparably across several language pairs. It permits automatic bias evaluation for translation from English to eight target languages with grammatical gender. The source side of WinoMT is 3888 concatenated sentences from Winogender \citep{rudinger-etal-2018-gender} and WinoBias \citep{zhao-etal-2018-gender}. These are coreference resolution datasets in which each sentence contains a primary entity which is co-referent with a pronoun -- \textit{the doctor} in the first example above and \textit{the developer} in the second -- and a secondary entity -- \textit{the nurse} and \textit{the cleaner} respectively. 

WinoMT evaluation extracts the grammatical gender of the primary entity from each translation hypothesis by automatic word alignment followed by morphological analysis. WinoMT then compares the translated primary entity with the gold gender, with the objective being a correctly gendered translation. The authors emphasise the following metrics over the challenge set:

\begin{itemize}
    \item \textbf{Accuracy} -- percentage of hypotheses with the correctly gendered primary entity.
    \item $\mathbf{\Delta G}$ -- difference in $F_1$ score between the set of sentences with masculine entities and the set with feminine entities.  
    \item $\mathbf{\Delta S}$ -- difference in accuracy between the set of sentences with pro-stereotypical (`pro') entities and those with anti-stereotypical (`anti') entities, as determined by \citet{zhao-etal-2018-gender} using US labour statistics. For example, the `pro' set contains male doctors and female nurses, while `anti' contains female doctors and male nurses.
\end{itemize}
Our main objective is increasing accuracy. We also report on $\Delta G$ and $\Delta S$ for ease of comparison to previous work. Ideally the absolute values of $\Delta G$ and $\Delta S$ should be close to 0. A high positive $\Delta G$ indicates that a model translates male entities better, while a high positive $\Delta S$ indicates that a model stereotypes male and female entities. Large negative values for $\Delta G$ and $\Delta S$, indicating a bias towards female or anti-stereotypical translation, are as undesirable as large positive values.

We note that $\Delta S$ can be significantly skewed by low-accuracy systems. A model generating male forms for most test sentences, stereotypical roles or not, will have very low $\Delta S$, since its pro- and anti-stereotypical class accuracy will both be about 50\%. Consequently in Appendix A we  report:
\begin{itemize}
    \item \textbf{M:F} -- ratio of hypotheses with male predictions to those with female predictions.
\end{itemize}
This should  be close to 1.0, since WinoMT balances male- and female-labelled sentences. M:F correlates strongly with $\Delta G$, but we consider M:F easier to interpret, particularly since very high or low M:F reduce the relevance of $\Delta S$.

Finally, we wish to reduce gender bias without reducing translation performance. We report \textbf{BLEU} \citep{papineni2002bleu} on separate, general test sets for each language pair. WinoMT is designed to work without target language references, and so it is not possible to measure translation performance on this set by measures such as BLEU.

\subsection{Gender debiased datasets}

\subsubsection{Handcrafted profession dataset}
Our hypothesis is that the absence of gender bias can be treated as a small domain for the purposes of NMT model adaptation. In this case a well-formed small dataset may give better results than attempts at debiasing the entire original dataset.

We therefore construct a tiny, trivial set of gender-balanced English sentences which we can easily translate into each target language. The sentences follow the template:
\begin{center}\textit{The $[$PROFESSION$]$ finished $[$his$|$her$]$ work.}\end{center}
We refer to this as the \textit{handcrafted} set\footnote{Handcrafted sets available at \url{https://github.com/DCSaunders/gender-debias}}. Each profession is from the list collected by \citet{prates2019assessing} from US labour statistics. We simplify this list by removing field-specific adjectives. For example, we have a single profession `engineer', as opposed to specifying industrial engineer, locomotive engineer, etc. In total we select 194 professions, giving just 388 sentences in a gender-balanced set. 

With manually translated masculine and feminine templates, we simply translate the masculine and feminine forms of each listed profession for each target language. In practice this translation is via an MT first-pass for speed, followed by manual checking, but given available lexicons this could be further automated. We note that the handcrafted sets contain no examples of coreference resolution and very little variety in terms of grammatical gender. A set of more complex sentences targeted at the coreference task might further improve WinoMT scores, but would be more difficult to produce for new languages.

We wish to distinguish between a model which improves gender translation, and one which improves its WinoMT scores simply by learning the vocabulary for previously unseen or uncommon professions. We therefore create a \textit{handcrafted no-overlap} set, removing source sentences with professions occurring in WinoMT to leave 216 sentences. We increase this set back to 388 examples with balanced adjective-based sentences in the same pattern, e.g. \textit{The tall $[$man$|$woman$]$  finished $[$his$|$her$]$ work}.

\subsubsection{Counterfactual datasets}
\begin{figure}[ht!]
\centering
\small
\includegraphics[width=1\linewidth]{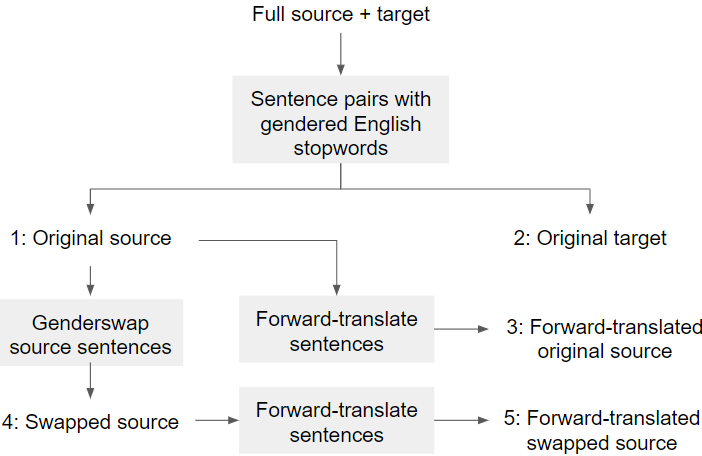}
\caption{Generating counterfactual datasets for adaptation. The \textbf{Original} set is 1$||$2, a simple subset of the full dataset. \textbf{FTrans original} is 1$||$3, \textbf{FTrans swapped} is 4$||$5, and \textbf{Balanced} is 1,4$||$2,5}
\label{fig:flowchart}

\end{figure}
\label{sec:counterfactual}
For contrast, we fine-tune on an approximated counterfactual dataset. Counterfactual data augmentation is an intuitive solution to bias from data over-representation \citep{lu2018gender}. It involves identifying the subset of sentences containing bias -- in this case gendered terms -- and, for each one, adding an equivalent sentence with the bias reversed -- in this case a gender-swapped version.

While counterfactual data augmentation is relatively simple for sentences in English, the process for inflected languages is challenging, involving identifying and updating words that are co-referent with all gendered entities in a sentence. Gender-swapping MT training data additionally requires that the same entities are swapped in the corresponding parallel sentence. A robust scheme for gender-swapping multiple entities in inflected language sentences directly, together with corresponding parallel text, is beyond the scope of this paper.  Instead we suggest a rough but straightforward approach for counterfactual data augmentation for NMT which to the best of our knowledge is the first application to parallel sentences. 

We first perform simple gender-swapping on the subset of the English source sentences with gendered terms. We use the approach described in \citet{zhao-etal-2018-gender} which swaps a fixed list of gendered stopwords  (e.g. \textit{man} / \textit{woman}, \textit{he} /  \textit{she}).\footnote{The stopword list and swapping script are provided by the authors of \citet{zhao-etal-2018-gender} at \url{https://github.com/uclanlp/corefBias}}. We then greedily forward-translate the gender-swapped English sentences with a baseline NMT model trained on the the full source and target text, producing gender-swapped target language sentences. 

This lets us compare four related sets for gender debiasing adaptation, as illustrated in Figure \ref{fig:flowchart}:
\begin{itemize}
    \item \textbf{Original}: a subset of parallel sentences from the original training data where the source sentence contains gendered stopwords.
    \item \textbf{Forward-translated (FTrans) original}: the source side of the \emph{original} set with forward-translated target sentences. 
    \item \textbf{Forward-translated (FTrans) swapped}: the \emph{original} source sentences are gender-swapped, then forward-translated to produce gender-swapped target sentences.
    \item \textbf{Balanced}: the concatenation of the \emph{original} and \emph{FTrans swapped} parallel datasets. This is twice the size of the other counterfactual sets.
\end{itemize}
 Comparing performance in adaptation of \textit{FTrans swapped} and \textit{FTrans original} lets us distinguish between the effects of gender-swapping and of obtaining target sentences from forward-translation.

\subsection{Debiasing while maintaining general translation performance}

Fine-tuning a converged neural network on data from a distinct domain typically leads to catastrophic forgetting of the original domain \citep{french1999catastrophic}. We wish to adapt to the gender-balanced domain without losing general translation performance. This is a particular problem when fine-tuning on the very small and distinct handcrafted adaptation sets.

\subsubsection{Regularized training}
Regularized training is a well-established approach for minimizing catastrophic forgetting during domain adaptation of machine translation \citep{miceli-barone-etal-2017-regularization}. One effective form is Elastic Weight Consolidation (EWC) \citep{kirkpatrick2017overcoming} which in NMT has been shown to maintain or even improve original domain performance \citep{thompson-etal-2019-overcoming, saunders-etal-2019-domain}. In EWC a regularization term is added to the original log likelihood loss function $L$ when training the debiased model (DB):
\begin{equation}
L'(\theta^{DB}) = L(\theta^{DB}) + \lambda \sum_j F_j (\theta^{DB}_j - \theta^{B}_{j})^2
\label{eq:regularization}
\end{equation}
$\theta^{B}_{j}$ are the converged parameters of the original biased model, and $\theta^{DB}_j$ are the current debiased model parameters. $F_j=\mathbb{E} \big[ \nabla^2  L(\theta^{B}_j)\big] $, a Fisher information estimate over samples from the biased data under the biased model. 
We apply EWC when performance on the original validation set drops, selecting hyperparameter $\lambda$ via validation set BLEU. 
\subsubsection{Gender-inflected search spaces for rescoring with debiased models}

\begin{figure}[th!]	
	\centering
	
	\begin{subfigure}[t]{\linewidth}
		\centering
		\includegraphics[width=1.6in]{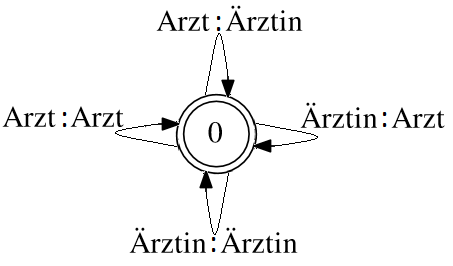}
		\caption{A subset of flower transducer $T$. $T$ maps vocabulary to itself as well as to differently-gendered inflections.}\label{fig:fsta}		
	\end{subfigure}

	\begin{subfigure}[t]{\linewidth}
		\centering
		\includegraphics[width=2in]{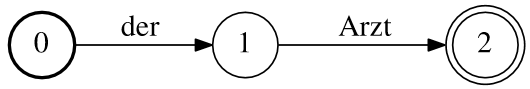}
		\caption{Acceptor $Y_B$ representing the biased first-pass translation $\mathbf{y_B}$ for source fragment 'the doctor'. The German hypothesis has the male form.}\label{fig:fstb}
	\end{subfigure}

	\begin{subfigure}[t]{\linewidth}
		\centering
		\includegraphics[width=2in]{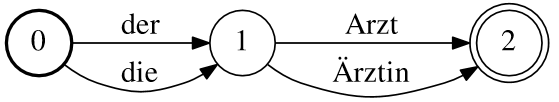}
		\caption{Gender-inflected search space constructed from the biased hypothesis `der Arzt'. Projection of the composition $Y_B\circ T$ contains paths with differently-gendered inflections of the original biased hypothesis. This lattice can now be rescored by a debiased model.}\label{fig:fstc}
	\end{subfigure}
	\caption{Finite State Transducers for lattice rescoring.}\label{fig:fst}
\end{figure}

An alternative approach for avoiding catastrophic forgetting takes inspiration from lattice rescoring for NMT \citep{stahlberg-etal-2016-syntactically} and Grammatical Error Correction \citep{stahlberg2019neural}. We assume we have two NMT models. With one we decode fluent translations which contain gender bias ($B$). For the one-best hypothesis we would translate:
    \begin{equation}
         \mathbf{y_B} = \text{argmax}_\mathbf{y} p_{B}(\mathbf{y} |\mathbf{x})
    \end{equation}

The other model has undergone debiasing ($DB$) at a cost to translation performance, producing:
    \begin{equation}
         \mathbf{y_{DB}} = \text{argmax}_\mathbf{y} p_{DB}(\mathbf{y} |\mathbf{x})
    \end{equation}
We construct a flower transducer $T$ that maps each word in the target language's vocabulary to itself, as well as to other forms of the same word with different gender inflections  (Figure \ref{fig:fsta}). We also construct $Y_B$, a lattice with one path representing the biased but fluent hypothesis $\mathbf{y_B}$ (Figure \ref{fig:fstb}).

The acceptor ${\mathcal P}(\mathbf{y_B}) = \text{proj}_\text{output} (Y_B \circ T )$ defines a language consisting of all the gender-inflected versions of the biased first-pass translation $\mathbf{y_B}$ that are allowed by $T$ (Figure \ref{fig:fstc}). We can now decode with lattice rescoring  ($LR$) by constraining inference to  ${\mathcal P}({\mathbf{y_B}})$:
    \begin{equation}
\mathbf{y_{LR}} = \text{argmax}_{ {\mathbf y} \in {\mathcal P}({\mathbf{y_B}}) } p_{DB}( {\mathbf y} | {\mathbf x})    \end{equation}

In practice we use beam search to decode the various hypotheses, and construct $T$ using heuristics on large vocabulary lists for each target language.

\section{Experiments}
\label{ss:data}
\subsection{Languages and data}
WinoMT provides an evaluation framework for translation from English to eight diverse languages. We select three pairs for experiments: English to German (en-de), English to Spanish (en-es) and English to Hebrew (en-he). Our selection covers three language groups with varying linguistic properties: Germanic, Romance and Semitic. Training data available for each language pair also varies in quantity and quality. We filter training data based on parallel sentence lengths and length ratios. 

For \textbf{en-de}, we use 17.6M sentence pairs from WMT19 news task datasets \citep{barrault-etal-2019-findings}. We validate on newstest17 and test on newstest18. 

For \textbf{en-es} we use 10M sentence pairs from the United Nations Parallel Corpus \citep{ziemski-etal-2016-united}. While still a large set, the UNCorpus exhibits far less diversity than the en-de training data. We validate on newstest12 and test on newstest13.

For \textbf{en-he} we use 185K sentence pairs from the multilingual TED talks corpus \citep{cettolo2014report}. This is both a specialized domain and a much smaller training set. We validate on the IWSLT 2012 test set and test on IWSLT 2014.

Table \ref{tab:data} summarises the sizes of datasets used, including their proportion of gendered sentences and ratio of sentences in the English source data containing male and female stopwords. A gendered sentence contains at least one English gendered stopword as used by \citet{zhao-etal-2018-gender}. 

Interestingly all three datasets have about the same proportion of gendered sentences: 11-12\% of the overall set. While en-es appears to have a much more balanced gender ratio than the other pairs, examining the data shows this stems largely from sections of the UNCorpus containing phrases like `empower women' and `violence against women', rather than gender-balanced professional entities.

\begin{table}[!ht]
\centering
\small 
\begin{tabular}{ |l|c |c| c|c| c|}
\hline 
 &\textbf{Training}  & \textbf{Gendered training}& \textbf{M:F}  &\textbf{Test}\\
\hline
en-de &  17.5M & 2.1M  & 2.4  & 3K\\
en-es & 10M & 1.1M &1.1 & 3K \\
en-he & 185K & 21.4K & 1.8 & 1K\\
\hline
    \end{tabular}
    \caption{Parallel sentence counts. A gendered sentence pair has minimum one gendered stopword on the English side. M:F is ratio of male vs female gendered training sentences.}
        \label{tab:data}
\end{table}

For en-de and en-es we learn joint 32K BPE vocabularies on the training data \citep{sennrich-etal-2016-neural}. For en-he we use separate source and target vocabularies. The Hebrew vocabulary is a 2k-merge BPE vocabulary, following the recommendations of \citet{ding-etal-2019-call} for smaller vocabularies when translating into lower-resource languages. 
For the en-he source vocabulary we experimented both with learning a new 32K vocabulary and with reusing the joint BPE vocabulary trained on the largest set – en-de – which lets us initialize the en-he system with the pre-trained en-de model. The latter resulted in higher BLEU and faster training.

\subsection{Training and inference}
For all models we use a Transformer model \citep{vaswani2017attention} with the `base' parameter settings given in Tensor2Tensor \citep{vaswani-etal-2018-tensor2tensor}. We train baselines to validation set BLEU convergence on one GPU, delaying gradient updates by factor 4 to simulate 4 GPUs \citep{saunders-etal-2018-multi}. During fine-tuning training is continued without learning rate resetting. Normal and lattice-constrained decoding is via SGNMT\footnote{\url{https://github.com/ucam-smt/sgnmt}} with beam size 4. BLEU scores are calculated for cased, detokenized output using SacreBLEU  \citep{post-2018-call}
\subsection{Lattice rescoring with debiased models}

For lattice rescoring we require a transducer $T$ containing gender-inflected forms of words in the target vocabulary. To obtain the vocabulary for German we use all unique words in the full target training dataset. For Spanish and Hebrew, which have smaller and less diverse training sets, we use 2018 OpenSubtitles word lists\footnote{Accessed Oct 2019 from \url{https://github.com/hermitdave/FrequencyWords/}}. We then use DEMorphy \citep{altinok2018demorphy} for German, spaCy \citep{honnibal2017spacy} for Spanish and the small set of gendered suffixes for Hebrew \citep{schwarzwald1982feminine} to approximately lemmatize each vocabulary word and generate its alternately-gendered forms. While there are almost certainly paths in $T$ containing non-words, we expect these to have low likelihood under the debiasing models. For lattice compositions we use the efficient OpenFST implementations \citep{allauzen2007openfst}. 

\subsection{Results}
\subsubsection{Baseline analysis}

\begin{table*}[!ht]
\centering
\small 

\noindent\makebox[\textwidth]{%
\begin{tabular}{ |l|ccc |ccc| ccc|}
\hline 
 &\multicolumn{3}{c|}{\textbf{en-de}}  &\multicolumn{3}{c|}{\textbf{en-es}}    &\multicolumn{3}{c|}{\textbf{en-he}}  \\
 & Acc & $\Delta G$ & $\Delta S$ & Acc & $\Delta G$ & $\Delta S$  & Acc & $\Delta G$ & $\Delta S$\\
 \hline

Microsoft & \textbf{74.1} & \textbf{0.0} & 30.2 & 47.3 & 36.8 & 23.2 & 48.1 & 14.9 & 32.9\\
Google & 59.4 & 12.5 & 12.5 & 53.1 & 23.4 & 21.3 & \textbf{53.7} & \textbf{7.9} & 37.8\\
Amazon & 62.4 & 12.9 & 16.7 & \textbf{59.4} & \textbf{15.4} & 22.3 & 50.5 & 10.3 & 47.3 \\
SYSTRAN & 48.6 & 34.5 & \textbf{10.3} & 45.6 & 46.3 & 15.0 & 46.6 & 20.5 & \textbf{24.5} \\
\hline
Baseline &  60.1 & 18.6 & 13.4 & 49.6 & 36.7 & \textbf{2.0} & 51.3 & 15.1 & 26.4\\
\hline
    \end{tabular}}
    \caption{WinoMT accuracy, masculine/feminine bias score $\Delta G$ and pro/anti stereotypical bias score $\Delta S$ for our baselines compared to commercial systems, whose scores are quoted directly from  \citet{stanovsky-etal-2019-evaluating}.}
        \label{tab:results-baseline}
\end{table*}
\begin{table*}[!ht]
\centering
\small 
\noindent\makebox[\textwidth]{%
\begin{tabular}{ |l|cccc |cccc| cccc|}
\hline 
 &\multicolumn{4}{c|}{\textbf{en-de}}  &\multicolumn{4}{c|}{\textbf{en-es}}    &\multicolumn{4}{c|}{\textbf{en-he}}  \\
 &  BLEU & Acc &  $\Delta G$ & $\Delta S$ &  BLEU & Acc&  $\Delta G$ & $\Delta S$ &  BLEU & Acc & $\Delta G$ & $\Delta S$\\
 \hline

Baseline & 42.7 & 60.1 &  18.6 & 13.4 & 27.8 & 49.6  & 36.7 & 2.0 & \textbf{23.8} & 51.3 & 15.1 & 26.4\\
\hline
Original & 41.8 & 60.7  & 15.9 & 15.6 & \textbf{28.3} & 53.0  & \textbf{24.3} & 10.8 & 23.5 & \textbf{53.6}  &\textbf{12.2} & 31.7\\
FTrans original & 43.3 & 60.0  & 20.0 & 13.9 & 27.4 &51.6&  31.6 & -4.8 & 23.4 & 48.7  & 23.0 & \textbf{20.9} \\
FTrans swapped & \textbf{43.4} & 63.0  & 15.4 & 12.7 & 27.4 & \textbf{53.7}  & 24.5 & -3.8 & 23.7 & 48.1  & 20.7 & 22.7\\
Balanced & 42.5 & \textbf{64.0}& \textbf{12.6}& \textbf{12.4} & 27.7 & 52.8& 26.2 & \textbf{1.9} & \textbf{23.8} & 48.3 & 20.8 & 24.0\\
\hline
    \end{tabular}}
    \caption{General test set BLEU and WinoMT scores after unregularised fine-tuning the baseline on four gender-based adaptation datasets. Improvements are inconsistent across language pairs.}
        \label{tab:results-counterfactual}
\end{table*}

In Table \ref{tab:results-baseline} we compare our three baselines to commercial systems on WinoMT, using results quoted directly from \citet{stanovsky-etal-2019-evaluating}. Our baselines achieve comparable accuracy, masculine/feminine bias score $\Delta G$ and pro/anti stereotypical bias score $\Delta S$ to four commercial translation systems, outscoring at least one system for each metric on each language pair. 

The $\Delta S$ for our en-es baseline is surprisingly small. Investigation shows this model predicts male and female entities in a ratio of over 6:1. Since almost all entities are translated as male, pro- and anti-stereotypical class accuracy are both about 50\%, making $\Delta S$ very small. This highlights the importance of considering  $\Delta S$ in the context of $\Delta G$ and M:F prediction ratio.

\subsubsection{Counterfactual adaptation}

Table \ref{tab:results-counterfactual} compares our baseline model with the results of unregularised fine-tuning on the counterfactual sets described in Section \ref{sec:counterfactual}.

Fine-tuning for one epoch on \textit{original}, a subset of the original data with gendered English stopwords, gives slight improvement in WinoMT accuracy and $\Delta G$ for all language pairs, while $\Delta S$ worsens. We suggest this set consolidates examples present in the full dataset, improving performance on gendered entities generally but emphasizing stereotypical roles.

On the \textit{FTrans original} set $\Delta G$ increases sharply relative to the \textit{original} set, while $\Delta S$ decreases. We suspect this set suffers from bias amplification \citep{zhao-etal-2017-men} introduced by the baseline system during forward-translation. The model therefore over-predicts male entities even more heavily than we would expect given the gender makeup of the adaptation data's source side. Over-predicting male entities lowers $\Delta S$ artificially. 

Adapting to \textit{FTrans swapped} increases accuracy and decreases both $\Delta G$ and  $\Delta S$ relative to the baseline for en-de and en-es. This is the desired result, but not a particularly strong one, and it is not replicated for en-he. The \textit{balanced} set has a very similar effect to the \textit{FTrans swapped} set, with a smaller test BLEU difference from the baseline.

We do find that the largest improvement in WinoMT accuracy consistently corresponds to the model predicting male and female entities in the closest ratio (see Appendix A). However, the best ratios for models adapted to these datasets are 2:1 or higher, and the accuracy improvement is small.

The purpose of EWC regularization is to avoid catastrophic forgetting of general translation ability. This does not occur in the counterfactual experiments, so we do not apply EWC. Moreover, WinoMT accuracy gains are small with standard fine-tuning, which allows maximum adaptation: we suspect EWC would prevent any improvements.

Overall, improvements from fine-tuning on counterfactual datasets  (\textit{FTrans swapped} and \textit{balanced}) are present. However, they are not very different from the improvements when fine-tuning on equivalent non-counterfactual sets (\textit{original} and \textit{FTrans original}). Improvements are also inconsistent across language pairs.

\subsubsection{Handcrafted profession set adaptation}
\begin{table*}[!ht]
\centering
\small 
\noindent\makebox[\textwidth]{%
\begin{tabular}{p{0.05cm}|p{2.3cm}|cccc |cccc| cccc|}
\cline{2-14} 
& &\multicolumn{4}{c|}{\textbf{en-de}}  &\multicolumn{4}{c|}{\textbf{en-es}}    &\multicolumn{4}{c|}{\textbf{en-he}}  \\
 & &  BLEU & Acc &  $\Delta G$ & $\Delta S$ &  BLEU & Acc& $\Delta G$ & $\Delta S$ &  BLEU & Acc &   $\Delta G$ & $\Delta S$\\
 \cline{2-14} 

\footnotesize{1}  & Baseline & \textbf{42.7} & 60.1  & 18.6 & 13.4 & \textbf{27.8} & 49.6  & 36.7 & 2.0 & 23.8 & 51.3  & 15.1 & 26.4\\
\footnotesize{2}  &  Balanced & 42.5 & 64.0 &12.6& 12.4 & 27.7 & 52.8& 26.2 & \textbf{1.9} & 23.8 & 48.3  & 20.8 & 24.0\\
\cline{2-14} 
\footnotesize{3} & Handcrafted (no overlap) &40.6 & 71.2 & 3.9 & 10.6 & 26.5 & 64.1 &  9.5 & -10.3 & 23.1 & 56.5 &  -6.2 & 28.9 \\

\footnotesize{4} & Handcrafted  & 40.8 & 78.3 & \textbf{-0.7} & 6.5 & 26.7 & 68.6 & 5.2 & -8.7 & 22.9 & 65.7 & -3.3 & 20.2\\

\cline{2-14} 
\footnotesize{5} & Handcrafted (converged) & 36.5 & \textbf{85.3} & -3.2 & 6.3 & 25.3 & \textbf{72.4} & \textbf{0.8} & -3.9 & 22.5 & \textbf{72.6}  & -4.2 & 21.0 \\

\footnotesize{6} & Handcrafted EWC &42.2 & 74.2  &2.2 & 8.4 & 27.2 & 67.8 & 5.8 & -8.2 & 23.3 & 65.2  & \textbf{-0.4} & 25.3 \\
\cline{2-14} 

\footnotesize{7} &Rescore 1 with 3 & \textbf{42.7} & 68.3  & 7.6 & 11.8  & \textbf{27.8} & 62.4  & 11.1 & -9.7  & \textbf{23.9}  & 56.2  & 2.8  &23.0  \\

\footnotesize{8} &Rescore 1 with 4 & \textbf{42.7} & 74.5 &  2.1 & 6.5  & \textbf{27.8} & 64.2 & 9.7 & -10.8  & \textbf{23.9}  & 58.4 & 2.7  &18.6  \\

\footnotesize{9} &Rescore 1 with 5& 42.5 & 81.7  & -2.4 & \textbf{1.5}  & 27.7 & 68.4  & 5.6 & -8.0  &  23.6 & 63.8& 0.7  & \textbf{12.9}  \\

\cline{2-14} 
    \end{tabular}}
    \caption{General test set BLEU and WinoMT scores after fine-tuning on the handcrafted profession set, compared to fine-tuning on the most consistent counterfactual set. Lines 1-2 duplicated from Table \ref{tab:results-counterfactual}. Lines 3-4 vary adaptation data. Lines 5-6 vary adaptation training procedure. Lines 7-9 apply lattice rescoring to baseline hypotheses.}
    \label{tab:handcrafted}
    \end{table*}

Results for fine-tuning on the handcrafted set are given in lines 3-6 of Table \ref{tab:handcrafted}. These experiments take place in minutes on a single GPU, compared to several hours when fine-tuning on the counterfactual sets and far longer if training from scratch. 

Fine-tuning on the handcrafted sets gives a much faster BLEU drop than fine-tuning on counterfactual sets. This is unsurprising since the handcrafted sets are domains of new sentences with consistent sentence length and structure. By contrast the counterfactual sets are less repetitive and close to subsets of the original training data, slowing forgetting. We believe the degradation here is limited only by the ease of fitting the small handcrafted sets.

Line 4 of Table \ref{tab:handcrafted} adapts to the handcrafted set, stopping when validation BLEU degrades by 5\% on each language pair. This gives a WinoMT accuracy up to 19 points above the baseline, far more improvement than the best counterfactual result. Difference in gender score $\Delta G$ improves by at least a factor of 4. Stereotyping score $\Delta S$ also improves far more than for counterfactual fine-tuning. Unlike the Table  \ref{tab:results-counterfactual} results, the improvement is consistent across all WinoMT metrics and all language pairs.

The model adapted to no-overlap handcrafted data (line 3) gives a similar drop in BLEU to the model in line 4. This model also gives stronger and more consistent WinoMT improvements over the baseline compared to the balanced counterfactual set, despite the implausibly strict scenario of no English profession vocabulary in common with the challenge set. This demonstrates that the adapted model does not simply memorise vocabulary.

The drop in BLEU and improvement on WinoMT can be explored by varying the training procedure. The model of line 5 simply adapts to handcrafted data for more iterations with no regularisation, to approximate loss convergence on the handcrafted set.  This leads to a severe drop in BLEU, but even higher WinoMT scores. 

In line 6 we regularise adaptation with EWC. There is a trade-off between general translation performance and WinoMT accuracy. With EWC regularization tuned to balance validation BLEU and WinoMT accuracy, the decrease is limited to about 0.5 BLEU on each language pair. Adapting to convergence, as in line 5, would lead to further WinoMT gains at the expense of BLEU.

 \begin{table*}[!ht]
\centering
\small 
\hspace{-2ex} 
\noindent\makebox[\textwidth]{%
\begin{tabular}{P{0.01cm}|P{1.38cm} P{1.33cm}P{1.22cm} |P{1.38cm} P{1.22cm}P{1.32cm} | P{1.38cm} P{1.33cm}P{1.38cm}|}
\cline{2-10} 
 &\multicolumn{3}{c|}{\textbf{en-de}}  &\multicolumn{3}{c|}{\textbf{en-es}}    &\multicolumn{3}{c|}{\textbf{en-he}}  \\
 & Acc & $\Delta G$ & $\Delta S$ & Acc & $\Delta G$ & $\Delta S$  & Acc & $\Delta G$ & $\Delta S$\\
\cline{2-10} 

\footnotesize{1}& \textbf{82.0} (74.1) & -3.0 (0.0) & 4.0 (30.2) & 65.8 (47.3) & 3.8 (36.8) & \textbf{1.9} (23.2) & 63.9 (48.1) & -2.6 (14.9) & 23.8 (32.9)\\
\footnotesize{2} & 80.0 (59.4) & -3.0 (12.5) & \textbf{2.7} (12.5) & 68.9 (53.1) & \textbf{0.6} (23.4) & 4.6 (21.3) & \textbf{64.6} (53.7) & -1.8 (7.9) & 21.5 (37.8)\\
\footnotesize{3} & 81.8 (62.4) & \textbf{-2.6} (12.9) & 4.3 (16.7) & \textbf{71.1} (59.4) & 0.7 (15.4) & 6.7 (22.3) & 62.8 (50.5) & \textbf{-1.1} (10.3) & 26.9 (47.3) \\
\footnotesize{4} & 78.4 (48.6) & -4.0 (34.5) & 5.3 (10.3) & 66.0 (45.6) & 4.2 (46.3) & -2.1 (15.0) & 62.5 (46.6) & -2.0 (20.5)  & \textbf{10.2} (24.5) \\
\cline{2-10} 
    \end{tabular}}
    \caption{We generate gender-inflected lattices from commercial system translations, collected by \citet{stanovsky-etal-2019-evaluating} (1: Microsoft, 2: Google, 3: Amazon, 4: SYSTRAN). We then rescore with the debiased model from line 5 of Table \ref{tab:handcrafted}. Scores are for the rescored hypotheses, with bracketed baseline scores duplicated from Table \ref{tab:results-baseline}.}
        \label{tab:commercial-rescore}
\end{table*}

\subsubsection{Lattice rescoring with debiased models}
In lines 7-9 of Table \ref{tab:handcrafted} we consider lattice-rescoring the baseline output, using three models debiased on the handcrafted data. 

Line 7 rescores the general test set hypotheses (line 1) with a model adapted to handcrafted data that has no source language profession vocabulary overlap with the test set (line 3). This scheme shows no BLEU degradation from the baseline on any language and in fact a slight improvement on en-he. Accuracy improvements on WinoMT are only slightly lower than for decoding with the rescoring model directly, as in line 3. 

In line 8, lattice rescoring with the non-converged model adapted to handcrafted data (line 4) likewise leaves general BLEU unchanged or slightly improved. When lattice rescoring the WinoMT challenge set, 79\%, 76\% and 49\% of the accuracy improvement is maintained on en-de, en-es and en-he respectively. This corresponds to accuracy gains of up to 30\% relative to the baselines with no general translation performance loss. 

In line 9, lattice-rescoring with the converged model of line 5 limits BLEU degradation to 0.2 BLEU on all languages, while maintaining 85\%, 82\% and 58\% of the WinoMT accuracy improvement from the converged model for the three language pairs. Lattice rescoring with this model gives accuracy improvements over the baseline of 36\%, 38\% and 24\% for en-de, en-es and en-he. 

Rescoring en-he maintains a much smaller proportion of WinoMT accuracy improvement than en-de and en-es. We believe this is because the en-he baseline is particularly weak, due to a small and non-diverse training set. The baseline must produce some inflection of the correct entity  before lattice rescoring can have an effect on gender bias.

\subsubsection{Reducing gender bias in `black box' commercial systems}
\label{sec:commercial}
Finally, in Table \ref{tab:commercial-rescore}, we apply the gender inflection transducer to the commercial system translations\footnote{The raw commercial system translations are provided by the authors of  \citet{stanovsky-etal-2019-evaluating} at \url{https://github.com/gabrielStanovsky/mt_gender}} listed in Table \ref{tab:results-baseline}. We find rescoring these lattices with our strongest debiasing model (line 5 of Table \ref{tab:handcrafted})  substantially improves WinoMT accuracy for all systems and language pairs.   

One interesting observation is that WinoMT accuracy after rescoring tends to fall in a fairly narrow range for each language relative to the performance range of the baseline systems.    For example, a 25.5\% range in baseline en-de accuracy becomes a 3.6\% range after rescoring. This suggests that our rescoring approach is not limited as much by the bias level of the baseline system as by the gender-inflection transducer and the models used in rescoring.     Indeed, we emphasise that the large improvements reported in Table \ref{tab:commercial-rescore} do not require any knowledge of the commercial systems or the data they were trained on; we use only the translation hypotheses they produce and our own rescoring model and transducer.

\section{Conclusions}
We treat the presence of gender bias in NMT systems as a domain adaptation problem. We demonstrate strong improvements under the WinoMT challenge set by adapting to tiny, handcrafted gender-balanced datasets for three language pairs. 

While naive domain adaptation leads to catastrophic forgetting, we further demonstrate two approaches to limit this: EWC and a lattice rescoring approach. Both allow debiasing while maintaining general translation performance. Lattice rescoring, although a two-step procedure, allows far more debiasing and potentially no degradation, without requiring access to the original model. 

We suggest small-domain adaptation as a more effective and efficient approach to debiasing machine translation than counterfactual data augmentation. We do not claim to fix the bias problem in NMT, but demonstrate that bias can be reduced without degradation in overall translation quality.
\section*{Acknowledgments}
This work was supported by EPSRC grants EP/M508007/1 and EP/N509620/1 and has been performed using resources provided by the Cambridge Tier-2 system operated by the University of Cambridge Research Computing Service\footnote{\url{http://www.hpc.cam.ac.uk}} funded by EPSRC Tier-2 capital grant EP/P020259/1.

\bibliographystyle{acl_natbib}
\bibliography{refs}

\appendix

\section{WinoMT male:female prediction ratio}
\begin{table*}[!ht]
\centering
\small 
\noindent\makebox[\textwidth]{%
\begin{tabular}{ |l|ccc |ccc| ccc|}
\hline 
 &\multicolumn{3}{c|}{\textbf{en-de}}  &\multicolumn{3}{c|}{\textbf{en-es}}    &\multicolumn{3}{c|}{\textbf{en-he}}  \\
 &  BLEU & Acc &  M:F &  BLEU & Acc& M:F & BLEU & Acc &  M:F \\
 \hline

Baseline & 42.7 & 60.1 &  3.4 & 27.8 & 49.6 & 6.3  & \textbf{23.8} & 51.3 & 2.2 \\
\hline
Original & 41.8 & 60.7 & 3.1 &  \textbf{28.3} & 53.0 & \textbf{4.0} & 23.5 & \textbf{53.6} & \textbf{2.0}\\
FTrans original & 43.3 & 60.0 & 3.9  & 27.4 &51.6& 5.4 & 23.4 & 48.7 & 3.0  \\
FTrans swapped & \textbf{43.4} & 63.0 & 3.1  & 27.4 & \textbf{53.7} & \textbf{4.0} & 23.7 & 48.1 & 2.6 \\
Balanced & 42.5 & \textbf{64.0}&\textbf{2.7}  & 27.7 & 52.8& 4.3& \textbf{23.8} & 48.3 & 2.7 \\
\hline
    \end{tabular}}
    \caption{General test set BLEU and WinoMT scores after unregularised fine-tuning the baseline on four gender-based adaptation datasets.}
        \label{tab:results-counterfactual-mf}
\end{table*}

\begin{table*}[!ht]
\centering
\small 
\noindent\makebox[\textwidth]{%
\begin{tabular}{ p{0.05cm}|l|ccc |ccc| ccc|}
\cline{2-11} 
 
 & &\multicolumn{3}{c|}{\textbf{en-de}}  &\multicolumn{3}{c|}{\textbf{en-es}}    &\multicolumn{3}{c|}{\textbf{en-he}}  \\
&  &  BLEU & Acc &  M:F &  BLEU & Acc& M:F & BLEU & Acc &  M:F \\
\cline{2-11} 

\footnotesize{1} &  Baseline & \textbf{42.7} & 60.1 &  3.4 & \textbf{27.8} & 49.6 & 6.3  & 23.8 & 51.3 & 2.2 \\
\footnotesize{2}  &  Balanced & 42.5 & 64.0&2.7  & 27.7 & 52.8& 4.3& 23.8 & 48.3 & 2.7 \\
\cline{2-11} 
\footnotesize{3} & Handcrafted (no overlap) &40.6 & 71.2 & 1.7 & 26.5 & 64.1 & 2.4 &   23.1 & 56.5 & 0.8\\

\footnotesize{4} & Handcrafted  & 40.8 & 78.3 &1.3 & 26.7 & 68.6 & 1.9&22.9 & 65.7 &0.9\\

\cline{2-11} 
\footnotesize{5} & Handcrafted (converged) & 36.5 & \textbf{85.3} & \textbf{0.9}& 25.3 & \textbf{72.4} & \textbf{1.5} & 22.5 & \textbf{72.6} &\textbf{1.0}\\

\footnotesize{6} & Handcrafted EWC &42.2 & 74.2 & 1.6& 27.2 & 67.8 & 2.0 & 23.3 & 65.2 & 1.2  \\
\cline{2-11} 

\footnotesize{7} &Rescore 1 with 3 & \textbf{42.7} & 68.3 & 2.2  & \textbf{27.8} & 62.4 & 2.3 &\textbf{23.9}  & 56.2 & 1.3   \\

\footnotesize{8} &Rescore 1 with 4 & \textbf{42.7} & 74.5 & 1.6 & \textbf{27.8} & 64.2 & 2.1   & \textbf{23.9}  & 58.4& 1.3  \\

\footnotesize{9} &Rescore 1 with 5& 42.5 & 81.7 & \textbf{1.1}  & 27.7 & 68.4 & 1.8 &  23.6 & 63.8& 1.3 \\

\cline{2-11} 
    \end{tabular}}
    \caption{General test set BLEU and WinoMT scores after fine-tuning on the handcrafted profession set, compared to fine-tuning on the most consistent counterfactual set. Lines 1-2 duplicated from Table \ref{tab:results-counterfactual-mf}. Lines 3-4 vary adaptation data. Lines 5-6 vary adaptation training procedure. Lines 7-9 apply lattice rescoring to baseline hypotheses.}
    \label{tab:handcrafted-mf}
    \end{table*}
We report $\Delta G$ on WinoMT for easy comparison to previous work, but also find that M:F prediction ratio on WinoMT is an intuitive and interesting metric. Tables \ref{tab:results-counterfactual-mf} and \ref{tab:handcrafted-mf} expand on the results of Tables \ref{tab:results-counterfactual} and \ref{tab:handcrafted} respectively.

\end{document}